\documentclass{article}

\usepackage[preprint,main]{neurips_2026}

\usepackage[utf8]{inputenc} 
\usepackage[T1]{fontenc}    
\usepackage{hyperref}       
\usepackage{url}            
\usepackage{booktabs}       
\usepackage{amsfonts}       
\usepackage{nicefrac}       
\usepackage{microtype}      

\usepackage{amsmath}
\usepackage{amssymb}
\usepackage[dvipsnames,table]{xcolor}

\usepackage{multirow}
\usepackage{graphicx}
\usepackage{float}
\usepackage{algorithm}
\usepackage{algorithmic}
\usepackage{wrapfig}

\title{Feat2Go: Visual Feature-Grounded Value Estimation for Embodied Reinforcement Learning}

\author{
  Junyang Shu \quad
  Zhiwei Lin \quad
  Bingqing Wei \quad
  Yongtao Wang\thanks{Corresponding author} \\
  Wangxuan Institute of Computer Technology, Peking University, China\\
  \texttt{\{jyshu25\}@stu.pku.edu.cn} \quad \texttt{\{wyt\}@pku.edu.cn} \\
}

\begin{document}

\maketitle
\begin{abstract}
Reinforcement learning is a promising approach for improving the capabilities of vision-language-action (VLA) models while avoiding the heavy data requirements of imitation learning.
However, its effectiveness for VLA models is often constrained by sparse supervision and the difficulty of designing informative reward signals for long-horizon manipulation.
In this work, we present Feat2Go, a fine-grained value estimation framework for embodied reinforcement learning.
Specifically, Feat2Go first derives a continuous progress target from a pretrained visual world model by measuring patch-level similarity to subgoal states and partitioning episodes into semantic stages with trend-based clustering.
We then train an embodied value model to predict this structural progress from the current observation and task instruction, and use the predicted value to reshape terminal rewards during policy optimization.
%
The proposed framework is compatible with existing VLA policy reinforcement learning pipelines, including PPO and GRPO, and does not rely on manual reward engineering.
%
Extensive experiments on ManiSkill3 and RoboTwin 2.0 demonstrate that Feat2Go consistently improves the performance of existing VLA models under both single-arm and bimanual manipulation settings.
More specifically, on ManiSkill3, Feat2Go improves OpenVLA-OFT from 17.5\% to 82.9\% average out-of-distribution success while retaining 96.9\% in-distribution performance.
On RoboTwin 2.0, Feat2Go achieves an average success rate of 88.8\% in domain-randomized task settings, outperforming prior reinforcement learning methods.
%
\end{abstract}

\begin{figure}[htbp]
    \centering
    \includegraphics[width=\textwidth]{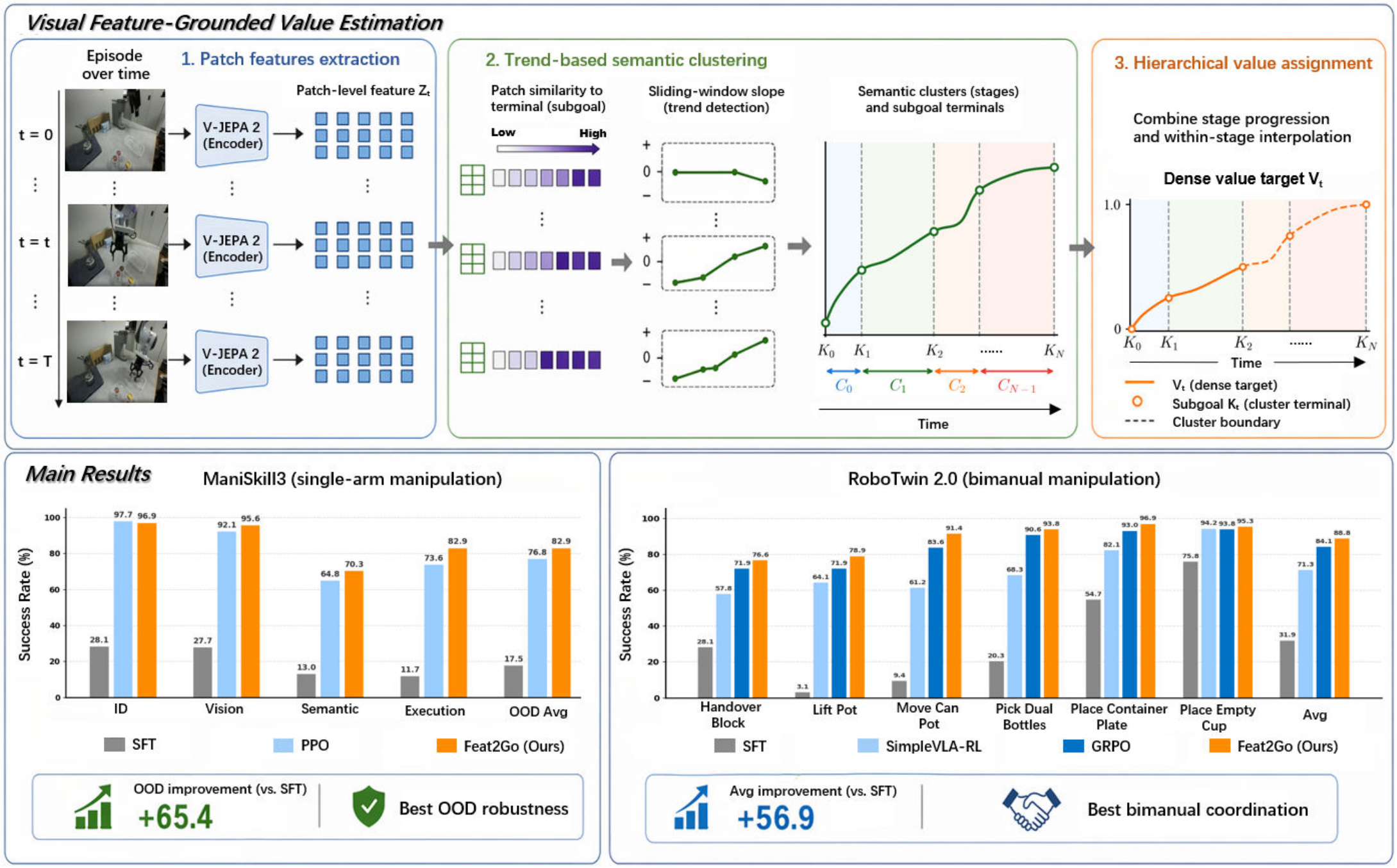}
    \vspace{-0.5cm}
    \caption{\textbf{Overview of Feat2Go value estimation and main results.} Feat2Go derives dense progress values from V-JEPA 2 features via trend-based semantic clustering and hierarchical value assignment, leading to strong VLA performance gains on ManiSkill3 and RoboTwin 2.0.}
    \label{fig:teaser}
    \vspace{-0.2cm}
\end{figure}

\section{Introduction}
\label{sec:introduction}
VLA models have recently emerged as a compelling paradigm for robotic manipulation, unifying visual perception, language conditioning, and action generation within a single policy interface~\citep{kim2024openvla,black2024pi0,black2025pi05}. 
By leveraging large-scale pretraining and supervised fine-tuning on robot demonstrations, these models have shown promising instruction-following and generalization capabilities across diverse manipulation settings~\citep{brohan2023rt2,kim2024openvla,black2024pi0}. 
However, this progress remains tightly coupled to increasingly large imitation-learning pipelines built on expensive human-collected data such as DROID~\citep{khazatsky2024droid}. 
Scaling such data collection is operationally demanding and time-consuming, particularly when extending to new environments or embodiments~\citep{khazatsky2024droid}. 
This makes reinforcement learning (RL) an attractive complement to imitation learning.
Instead of relying solely on ever-larger demonstration corpora, RL offers a mechanism to improve VLA policies through interaction, thereby enhancing capability while reducing dependence on massive labeled robot data~\citep{liu2025whatrlbring,guo2025irevla,simplevlarl2025,rlinfvla2025}.

Despite this promise, applying RL to VLA policies remains challenging, primarily due to the severe bottleneck of sparse rewards.
In embodied control, supervision is typically restricted to binary episode outcomes (\textit{e.g.}, 0 for failure and 1 for success)~\citep{andrychowicz2017her,arjonamedina2019rudder}. 
This sparsity significantly obfuscates long-horizon credit assignment, as the actual degree of task progression can vary drastically even among uniformly ``failed'' episodes.
While shaping dense, proxy rewards could theoretically alleviate this issue~\citep{ng1999policyinvariance}, manually engineering rule-based progress metrics for topologically diverse and physically complex tasks is prohibitively difficult and inherently non-scalable~\citep{andrychowicz2017her}.
Moreover, recent RL frameworks~\citep{guo2025irevla,simplevlarl2025,rlinfvla2025,liu2025whatrlbring} for VLA show that there is a critical supervisory interface gap between pretrained VLA representations and downstream policy optimization.
These observations motivate a representation-driven alternative: instead of manually designing rewards or relying on coarse heuristic progress proxies, can we extract a dense and semantically meaningful progress signal directly from complete embodied episodes?

%
To this end, this paper introduces \emph{Feat2Go}, a fine-grained value estimation framework for embodied RL. 
%
The core of Feat2Go is to construct a feature-grounded notion of task progress by leveraging a pretrained visual world model and completed episodes with explicit initial and terminal states.
Concretely, we first encode episode frames with V-JEPA 2~\citep{assran2025vjepa2}, compare intermediate states to subgoal or terminal states in representation space, and partition each episode into semantic stages through trend-based clustering. 
This procedure yields a dense structural value target that captures both stage-level progression and inter-stage refinement. 
We then train an embodied value model to predict this progress signal from the current observation and task instruction, and use the predicted structural value to reshape terminal rewards during policy optimization. 
%
The overall framework is compatible with both PPO and GRPO~\citep{schulman2017ppo,shao2024deepseekmath} and provides a value-estimation target without a handcrafted reward function, as summarized in Figure~\ref{fig:teaser}.

We evaluate Feat2Go on ManiSkill3~\citep{tao2024maniskill3} for single-arm manipulation and RoboTwin 2.0~\citep{chen2025robotwin2} for bimanual manipulation under challenging out-of-distribution and domain-randomized settings. 
The results show that the proposed feature-grounded value estimation substantially improves robust policy learning across both settings. 
Empirically, our method yields substantial gains in generalization: on ManiSkill3, Feat2Go boosts the out-of-distribution success rate of OpenVLA-OFT by an absolute margin of 65.4\%, while strictly preserving its in-distribution proficiency (96.9\%). 
Furthermore, even under the severe domain randomizations of RoboTwin 2.0, Feat2Go demonstrates remarkable robustness by reaching an 88.8\% success rate, consistently outperforming established RL baselines. 
%
These results suggest that translating pretrained visual structure into progress-aware RL supervision is an effective way to improve the robustness, generalization, and quality of policy learning in VLA models.

In summary, our contributions are threefold:
\begin{itemize}
    \item We introduce a fine-grained value estimation framework that derives structural progress targets from a pretrained visual world model and semantic episode partitioning.
    \item We develop an embodied value model that predicts progress from the current observation and task instruction, and integrate it with PPO and GRPO through reward reshaping for VLA reinforcement learning.
    \item We demonstrate strong empirical gains on ManiSkill3 and RoboTwin 2.0, showing that Feat2Go improves policy capability and generalization across both single-arm and bimanual manipulation settings.
\end{itemize}

\section{Related Work}
\label{sec:related_work}

\subsection{VLA for Robotic Manipulation}
Recent years have seen rapid progress in VLA models that unify perception, language grounding, and action prediction for end-to-end robotic control. 
Early large-scale policy architectures such as RT-1 and RT-2 demonstrated that increasing model capacity and leveraging internet-scale vision-language priors can substantially improve robotic generalization~\citep{brohan2022rt1,brohan2023rt2}. 
Subsequent efforts expanded both the scale and openness of this paradigm through large multi-robot datasets and open policy training pipelines, including Open X-Embodiment, DROID, and OpenVLA~\citep{openx2023,khazatsky2024droid,kim2024openvla}. 
More recent foundation policies such as $\pi_0$ and $\pi_{0.5}$ further strengthened zero-shot transfer and open-world manipulation capabilities~\citep{black2024pi0,black2025pi05}. 
These advances establish VLA models as a powerful backbone for embodied manipulation, but they also remain strongly tied to large-scale supervised imitation learning on human-collected demonstrations~\citep{openx2023,khazatsky2024droid,kim2024openvla}. 
Rather than proposing a new VLA policy architecture, this work focuses on improving downstream RL adaptation of VLA policies through a more effective progress-aware supervision signal.

%
\subsection{Reinforcement Learning for VLA}
Alongside the growth of VLA policies, a recent line of work has explored reinforcement learning to further improve policy capability and generalization beyond supervised fine-tuning~\citep{guo2025irevla,simplevlarl2025,rlinfvla2025,liu2025whatrlbring}. 
Existing studies have shown that RL can be effective for VLA adaptation, both through algorithmic refinements and improved training systems, including iterative RL-SFT schemes, integrated RL frameworks, and large-scale VLA RL infrastructure~\citep{guo2025irevla,simplevlarl2025,rlinfvla2025}. 
Much empirical evidence suggests that RL is particularly helpful for semantic and execution generalization in VLA settings~\citep{liu2025whatrlbring}. 
%
More recently, some works have begun to incorporate embodied value models, critics, or self-referential feedback mechanisms into VLA improvement pipelines~\citep{shu2025rftf,zhai2025vlac,fei2025srpo}. 
Self-Improving Embodied Foundation Models similarly advocate a two-stage post-training pipeline in which a learned progress-related objective (Steps-To-Go) supports autonomous online improvement~\citep{ghasemipour2025selfimproving}. 
\(\pi^{*}_{0.6}\) and its RECAP training recipe show that VLA models can learn from experience by combining offline RL, on-policy data, and advantage-conditioned adaptation during deployment~\citep{amin2025pi06}. 
Though these works strongly support the importance of learned feedback signals for VLA improvement, they parameterize feedback in different ways, such as temporal objectives, progress deltas, or self-referential rewards, rather than deriving an explicitly stage-structured value target from complete episodes~\citep{shu2025rftf,zhai2025vlac,ghasemipour2025selfimproving,fei2025srpo}. 
In contrast, the proposed Feat2Go constructs its supervision from a pretrained visual world model and completed episodes, using representation-space similarity and semantic episode partitioning to derive a fine-grained structural value target. 
This positions our method as a representation-driven value estimation framework that complements recent RL-for-VLA advances while providing a more explicit interface between pretrained visual structure and downstream reward reshaping.

\section{Method}
\label{sec:method}
\subsection{Fine-Grained Value Estimation}
\label{sec:3.1}
A recent line of embodied value modeling uses ``Steps-To-Go''-style objectives, wherein the value of a given frame is determined by the number of time steps remaining until task completion~\citep{ghasemipour2025selfimproving, amin2025pi06}. 
However, this formulation overlooks the reality that the contribution of different actions to overall task progress varies significantly.
Consequently, such metrics struggle to provide the fine-grained supervisory signals necessary for robust value estimation. 

To overcome the aforementioned limitation, we propose quantifying the progression of embodied tasks directly at the feature representation level. 
We first use the V-JEPA 2 encoder~\citep{assran2025vjepa2} to process each frame of the embodied episode, transforming raw video tensors with dimensions $[T, C, H, W]$ into a sequence of patch features with dimensions $[T, P, D]$, where $T$, $C$, $H$, $W$, $P$, and $D$ denote the episode length, channels, height, width, number of patches, and feature dimension, respectively.
%
To quantify the feature-level similarity between any given frame and the terminal frame, we then compute a patch-wise composite metric.
Specifically, for each spatial patch, this metric captures differences in both vector orientation and magnitude by multiplying the cosine similarity by the natural exponential of the $\beta$-regularized Manhattan distance. 
%
The overall frame-level similarity $S$ is then obtained by averaging these scores across all patches:
\begin{equation}
    S(\mathbf{Z}_t, \mathbf{Z}_{K}) = \frac{1}{P} \sum_{p=1}^{P} \left( \frac{\mathbf{z}_{t,p} \cdot \mathbf{z}_{K,p}}{\|\mathbf{z}_{t,p}\|_2 \|\mathbf{z}_{K,p}\|_2} \cdot \exp \left( -\beta \|\mathbf{z}_{t,p} - \mathbf{z}_{K,p}\|_1 \right) \right),
    \label{eq:similarity}
\end{equation}
where $\mathbf{Z}_t, \mathbf{Z}_{K} \in \mathbb{R}^{P \times D}$ denote the feature tensors of a given frame at time step $t$ and the terminal frame $K$, and $\mathbf{z}_{t,p}, \mathbf{z}_{K,p} \in \mathbb{R}^D$ represent their respective $p$-th patch feature vectors.
$\beta > 0$ is a scaling hyperparameter controlling the sensitivity to the magnitude difference.

After that, we apply a trend mutation detection algorithm based on sliding windows and linear fitting to partition the embodied episode into semantic clusters.
Specifically, given a predefined maximum limit for frames per cluster $fpc_{\max}$ 
and a fixed window size $W$, the algorithm proceeds iteratively backward from the unassigned terminal frame $K$. 
At each step, we initiate the window search at index $t = K - fpc_{\max} + 1$ and perform linear regression on a short window of length $W$ extending forward from the candidate frame $t$. Crucially, the similarity score is computed relative to the current unassigned terminal frame $K$, rather than the global terminal frame of the episode. 

Therefore, the independent variable is the relative frame index $\tau \in \{0, 1, \dots, W-1\}$, and the dependent variable is the local similarity score $y_\tau = S(\mathbf{Z}_{t+\tau}, \mathbf{Z}_K)$.
The optimal slope $w$ can be computed in closed form as:
\begin{equation}
    w = \frac{\sum_{\tau=0}^{W-1} (\tau - \bar{\tau})(y_\tau - \bar{y})}{\sum_{\tau=0}^{W-1} (\tau - \bar{\tau})^2},
\end{equation}
where $\bar{\tau}$ and $\bar{y}$ denote the means of the indices and corresponding similarity scores within the window, respectively.
If the estimated slope exceeds a designated mutation threshold hyperparameter $\alpha$ (\textit{i.e.}, $w > \alpha$), the frame at index $t$ is identified as a critical transition point. 
This thresholding identifies the shift from a ``noise-dominated'' region to a ``trend-dominated'' regime. 
Once a cluster boundary is confirmed, the segment from $t$ to $K$ forms a new cluster, and we update $K = t - 1$ to repeat the entire process until the entire episode is partitioned.

Once the episode is sequentially partitioned into $N$ semantic clusters (indexed from $n = 0$ to $N-1$), the structural value assignment must harmoniously fuse macroscopic stage-wise progression with microscopic frame-level dynamics. 
To achieve this, we formulate the integrated value $V_t$ for any state $t$ via a hierarchical interpolation scheme:
\begin{equation}
    V_t = \underbrace{\frac{1}{N} \sum_{i=0}^{N-1} i \cdot \mathbb{I}(t \in \mathcal{C}_i)}_{\text{Macroscopic Progression}} + \underbrace{\vphantom{\sum_{i=0}^{N-1}} \frac{1}{N} \cdot S_{norm}(\mathbf{Z}_t, \mathbf{Z}_{K_n})}_{\text{Microscopic Interpolation}}, \quad \forall t \in \mathcal{C}_n
\end{equation}
where $\mathcal{C}_n$ denotes the index set of the $n$-th temporal cluster, and $\mathbb{I}(\cdot)$ is the indicator function. 
Intuitively, the macroscopic term captures coarse task progress by tracking the number of completed temporal stages. 
The microscopic term then provides continuous, dense guidance within each stage based on $S_{norm}(\mathbf{Z}_t, \mathbf{Z}_{K_n}) \in [0, 1]$, which represents the min-max normalized feature similarity between the current state $t$ and the local sub-goal $K_n$. 
%
This dual-grained composition ensures a monotonically increasing, fine-grained value landscape throughout the embodied objective.

\begin{algorithm}[htpb]
\caption{Fine-Grained Value Estimation}
\label{alg:value_estimation}
\begin{algorithmic}[1]
\REQUIRE Raw episode frames $\mathcal{V} = \{v_0, \dots, v_{T-1}\}$, V-JEPA 2 encoder, scaling factor $\beta$, window size $W$, max frames per cluster $fpc_{\max}$, mutation threshold $\alpha$.
\ENSURE Integrated structural value $V_t$ for all states $t \in \{0, \dots, T-1\}$.

\STATE \textbf{/* Phase 1 \& 2: Feature Extraction \& Semantic Clustering */}
\STATE Extract patch features $\mathbf{Z}_t \leftarrow \text{V-JEPA 2}(v_t)$ for all $t$ \hfill \COMMENT{$\mathbf{Z}_t \in \mathbb{R}^{P \times D}$}
\STATE Initialize unassigned terminal frame $K \leftarrow T - 1$, and empty list $\text{Clusters}$
\WHILE{$K \ge 0$}
    \STATE $\text{cut} \leftarrow \max(0, K - fpc_{\max} + 1)$ \hfill \COMMENT{Default cut-off fallback}
    \FOR{candidate frame $t = \text{cut}$ \TO $K - W$}
        \STATE Compute slope $w$ via linear regression on $\{(\tau, S(\mathbf{Z}_{t+\tau}, \mathbf{Z}_K))\}_{\tau=0}^{W-1}$ (Eq.~\ref{eq:similarity})
        \IF{$w > \alpha$}
            \STATE $\text{cut} \leftarrow t$
            \STATE \textbf{break} \hfill \COMMENT{Critical transition detected}
        \ENDIF
    \ENDFOR
    \STATE $\text{Clusters.prepend}(\{\text{cut}, \dots, K\})$ \hfill \COMMENT{Prepend to maintain chrono-order}
    \STATE $K \leftarrow \text{cut} - 1$
\ENDWHILE

\STATE \textbf{/* Phase 3: Hierarchical Value Assignment */}
\STATE Let the chronological clusters be $\{\mathcal{C}_0, \mathcal{C}_1, \dots, \mathcal{C}_{N-1}\}$
\FOR{$n = 0$ \TO $N-1$}
    \STATE Sub-goal terminal state $K_n \leftarrow \max(\mathcal{C}_n)$ 
    \FOR{each $t \in \mathcal{C}_n$}
        \STATE $V_t \leftarrow \frac{1}{N} \sum_{i=0}^{N-1} i \cdot \mathbb{I}(t \in \mathcal{C}_i) + \frac{1}{N} \cdot S_{norm}(\mathbf{Z}_t, \mathbf{Z}_{K_n})$
    \ENDFOR
\ENDFOR

\RETURN Trajectory value sequence $\{V_0, V_1, \dots, V_{T-1}\}$
\end{algorithmic}
\end{algorithm}

We build our embodied value model upon the Qwen3-VL-4B-Instruct architecture to leverage its robust vision-language reasoning capabilities.
The model takes a tripartite input, \textit{i.e.}, a system prompt defining the judge's role, the current visual frame, and the embodied task instruction. 
%
Due to autoregressive modeling, the last generated token possesses a global receptive field. 
Consequently, its hidden state naturally encodes a comprehensive fusion of both visual and textual context. 
%
Thus, we extract the last hidden representation and use it to predict the value through an MLP-based value head. 
Moreover, inspired by $\pi_{0.6}^*$~\citep{amin2025pi06}, we convert the continuous value estimation into a discrete prediction problem. 
Specifically, we uniformly discretize the continuous target value space $[0, 1]$ into $B = 201$ distinct bins. 
%
%
This discrete binning formulation bypasses the instability often associated with direct scalar regression.
Furthermore, to mitigate distribution shifts between the pre-trained VLM and the downstream value estimation task, and to ensure training stability, we incorporate Layer Normalization and GELU activation layers within the value head.

\subsection{Integration into Reinforcement Learning}
\label{sec:3.2}
%

\begin{figure}[t]
    \centering
    \includegraphics[width=\linewidth]{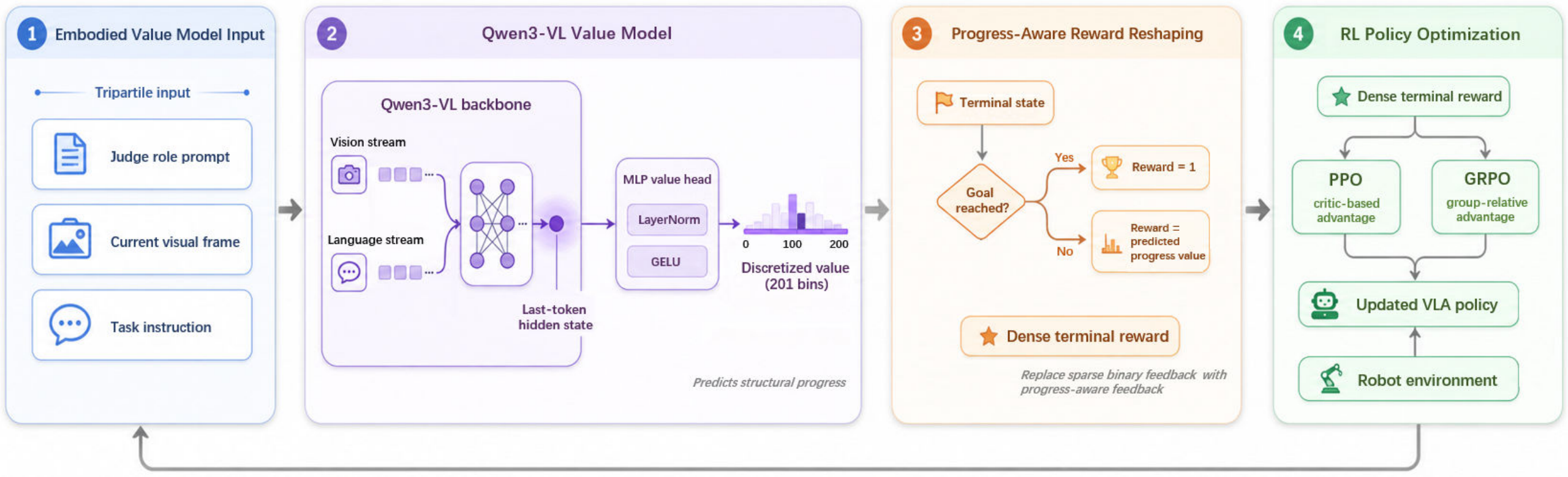}
    \caption{Feat2Go trains an embodied value model from visual-language inputs to predict a discretized progress value, which is used to reshape sparse terminal rewards. The resulting dense terminal reward is then integrated into PPO or GRPO to provide progress-aware policy optimization.}
    \label{fig:integration_into_rl}
\end{figure}


%

To empirically validate the structural generalizability of our approach, as illustrated in Figure~\ref{fig:integration_into_rl}, we integrate the proposed embodied value model into two canonical reinforcement learning paradigms, \textit{i.e.}, PPO and GRPO~\citep{schulman2017ppo,shao2024deepseekmath}.
%
%
%
Specifically, let $\mathcal{\tau} = (\mathbf{s}_0, \mathbf{a}_0, \mathbf{r}_1, \dots, \mathbf{r}_T, \mathbf{s}_T)$ denote an episode of length $T$, where $\mathbf{s}_T$ denotes the terminal state. 
Traditional sparse reward formulations assign a naive terminal reward $R_{sparse}(\mathbf{s}_T) = \mathbb{I}(\mathbf{s}_T \in \mathcal{S}_{goal})$, where $\mathcal{S}_{goal}$ represents the set of successful terminal states, and $\mathbb{I}(\cdot)$ is the binary indicator function. 
The naive sparse reward yields zero for any sub-optimal episode and critically suppresses the learning signal.
%
To address this, we deploy our parameterized embodied value model $V_\theta(\cdot)$ as a progress-aware reward reshaper. 
Specifically, we formulate the augmented structural reward function $R_{dense}(\mathbf{s}_T)$ via a convex combination driven by the goal-condition indicator:
\begin{equation}
    R_{dense}(\mathbf{s}_T) = \mathbb{I}(\mathbf{s}_T \in \mathcal{S}_{goal}) + \Big( 1 - \mathbb{I}(\mathbf{s}_T \in \mathcal{S}_{goal}) \Big) \cdot V_\theta(\mathbf{s}_T),
    \label{eq:reward_shaping}
\end{equation}
where $V_\theta(\mathbf{s}_T) \in [0, 1]$ represents the value for failed attempts.
Then, we propagate this dense reward signal across the trajectory to formulate the state-action advantages. 

%
%
\textbf{PPO\ \ } 
The advantage function $\hat{A}_t$ of PPO is estimated via Generalized Advantage Estimation (GAE)~\citep{schulman2015gae} using a learned critic network $V_\phi$. 
Specifically, the GAE is formulated as an exponentially weighted average of temporal-difference errors:
\begin{equation}
    \hat{A}_t = \sum_{l=1}^{T-t} (\gamma \lambda)^{l-1} \delta_{t+l}, \quad \text{where} \quad \delta_{t+l} = r_{t+l} + \gamma V_\phi(\mathbf{s}_{t+l}) - V_\phi(\mathbf{s}_{t+l-1}),
    \label{eq:gae}
\end{equation}
where the intermediate step rewards are zero (\textit{i.e.}, $r_t = 0$ for $t < T$), and the dense structural evaluation is solely assigned at the terminal step ($r_T = R_{dense}(\mathbf{s}_T)$). 
Here, $\gamma \in (0, 1]$ is the discount factor, and $\lambda \in [0, 1]$ is the GAE bias-variance trade-off parameter.
%
Then, the policy actor $\pi_\omega$ is optimized via the standard clipped surrogate objective:
\begin{equation}
    \mathcal{L}_{\text{PPO}}(\omega) = - \mathbb{E}_{\tau} \left[ \sum_{t=0}^{T-1} \min \Big( \rho_t(\omega) \hat{A}_t, \; \text{clip}\big(\rho_t(\omega), 1 - \epsilon_{\text{low}}, 1 + \epsilon_{\text{high}}\big) \hat{A}_t \Big) \right],
    \label{eq:ppo_loss}
\end{equation}
where $\rho_t(\omega) = \frac{\pi_\omega(\mathbf{a}_t|\mathbf{s}_t)}{\pi_{\omega_{\text{old}}}(\mathbf{a}_t|\mathbf{s}_t)}$ is the importance sampling ratio. $\epsilon_{\text{low}}$ and $\epsilon_{\text{high}}$ denote the thresholds.
%
%

\textbf{GRPO\ \ } 
GRPO estimates advantages purely through group-relative distribution modeling. 
Specifically, during rollout, we sample a local group of $G$ trajectories $\{ \tau^{(i)} \}_{i=1}^G$. 
For the $i$-th trajectory in the group, its group-relative advantage $\tilde{A}^{(i)}$ is computed via intra-group standardization of our dense terminal rewards:
\begin{equation}
    \tilde{A}^{(i)} = \frac{R_{dense}^{(i)} - \mu}{\sigma + \delta},
\end{equation}
where $\mu$ and $\sigma$ represent the empirical mean and standard deviation of the terminal rewards $\{ R_{dense}^{(i)} \}_{i=1}^G$ within the sampled group. 
By propagating this standardized return backward through time, the step-wise advantage for the $i$-th trajectory is formulated as $\hat{A}_t^{(i)} = \gamma^{T-t}\tilde{A}^{(i)}$, where $\gamma \in (0, 1]$ is the discount factor.
The GRPO policy actor $\pi_\omega$ is subsequently optimized via a group-aggregated surrogate objective:
\begin{equation}
    \mathcal{L}_{\text{GRPO}}(\omega) = - \mathbb{E}_{\{\tau^{(i)}\}_{i=1}^G} \left[ \frac{1}{G} \sum_{i=1}^G \sum_{t=0}^{T-1} \min \Big( \rho_t^{(i)}(\omega) \hat{A}_t^{(i)}, \; \text{clip}\big(\rho_t^{(i)}(\omega), 1 - \epsilon_{\text{low}}, 1 + \epsilon_{\text{high}}\big) \hat{A}_t^{(i)} \Big) \right],
    \label{eq:grpo_loss}
\end{equation}
where $\rho_t^{(i)}(\omega) = \frac{\pi_\omega(\mathbf{a}_t^{(i)}|\mathbf{s}_t^{(i)})}{\pi_{\omega_{\text{old}}}(\mathbf{a}_t^{(i)}|\mathbf{s}_t^{(i)})}$ denotes the trajectory-specific importance sampling ratio, bounded by the asymmetric clipping ratios $\epsilon_{\text{low}}$ and $\epsilon_{\text{high}}$.

By substituting the structure-agnostic binary outcomes with our fine-grained value evaluation, the agent seamlessly acquires dense spatial-temporal gradients globally. 
This addresses the credit assignment challenge inherent in sparse-reward embodied reinforcement learning by providing dense, progress-aware feedback.

\section{Experiments}
\label{sec:experiments}
\subsection{Experimental Setup}
\textbf{Benchmarks\ \ }
To comprehensively evaluate our proposed method, we conduct experiments on two widely adopted simulation benchmarks: ManiSkill3~\citep{tao2024maniskill3} for single-arm manipulation and RoboTwin 2.0~\citep{chen2025robotwin2} for bimanual manipulation. 

\textit{ManiSkill3.}
Following the setup established by RL4VLA and RLinf-VLA~\citep{liu2025whatrlbring,rlinfvla2025}, we evaluate our method on standard pick-and-place tasks using an 8-DoF WidowX-250S robotic arm in ManiSkill3. 
To rigorously assess generalization capabilities, we evaluate performance across three distinct dimensions: 
(1) \textit{Vision}: evaluates visual robustness against dynamic textures, unseen table surfaces, and image-level noise; 
(2) \textit{Semantics}: targets compositional and linguistic generalization by introducing unseen objects and receptacles, paraphrased instructions, and multi-object/distractor scenarios; 
(3) \textit{Execution}: tests robustness to dynamic spatial variations, including randomized initial poses and mid-episode object repositioning.

\textit{RoboTwin 2.0.}
We configure the model with a pair of Agilex Piper robotic arms and assess its performance across six representative tasks. 
To systematically measure robust generalization, we adopt the benchmark's domain-randomized task settings, which leverage comprehensive domain randomization, including dynamic lighting, varied background textures, distractor objects, and diverse language instructions to significantly broaden the evaluation distribution~\citep{chen2025robotwin2}. 

To ensure a fair and direct comparison with RLinf-VLA, we strictly adopt their extensive environmental evaluation protocol~\citep{rlinfvla2025}. 
Specifically, we evaluate each task using 256 parallel environments in ManiSkill3, and across 128 unique environment configurations in RoboTwin 2.0 with diverse random seeds. 
This rigorous evaluation setup effectively mitigates the impact of experimental noise and random variance, leading to more statistically reliable results.


\textbf{Implementation Details\ \ }
Feat2Go has two stages: embodied value model training and reinforcement learning process.

\textit{Embodied Value Model.} 
For the network architecture, the weights of the value head are orthogonally initialized with all bias terms set to zero. 
The pre-training regime on the large-scale DROID dataset consists of two sequential stages: we first freeze the VLM backbone to train only the value head for 1 epoch, and subsequently unfreeze the backbone to jointly optimize the entire model for 12 epochs. 
For state-value representation learning, we set the window size $W=10$ and $fpc_{max} = 50$, with a mutation threshold $\alpha = 3 \times 10^{-3}$ and a scaling factor $\beta = 0.99$. 
During task-specific adaptation, we efficiently align the value model with the target environment's unique semantics and dynamics via lightweight fine-tuning. 
This process leverages a small curriculum of 1K successful episodes per task collected by an existing policy model, fine-tuning the embodied value model for $2,000$ gradient steps per task.
Training is batched by reading 16 episodes concurrently with 4 NVIDIA A100 GPUs, uniformly sampling $1,024$ image-value pairs per iteration. 
The learning rate is initialized to $1 \times 10^{-5}$ for pre-training and decayed to $1 \times 10^{-6}$ for downstream value fine-tuning.

\textit{Reinforcement Learning.} We employ LoRA parameter-efficient learning with a rank parameter of $32$. 
The optimization process lasts for 1,000 global steps. 
The objective uses a clipped surrogate formulation with bounds $[\epsilon_{\text{low}}, \epsilon_{\text{high}}] = [0.2, 0.28]$. 
Value estimation uses GAE with $\lambda = 0.95$ and a temporal discount factor $\gamma = 0.99$. 
Notably, given the substantial performance divergence between RL induction and the supervised behavioral cloning baseline, we deliberately set the KL penalty coefficient to $0$. 
We use 8 NVIDIA A100 GPUs for RL training.
For ManiSkill3, we use a global batch size of $640$, an actor learning rate of $1 \times 10^{-4}$, and a critic learning rate of $3 \times 10^{-3}$; for RoboTwin 2.0, recognizing that GRPO omits the explicit critic architecture, we configure a global batch size of $1,024$ and an actor learning rate of $5 \times 10^{-5}$.

\subsection{Main Results}
Our primary evaluation aims to answer whether the proposed Feat2Go can consistently enhance the robustness and OOD generalization of VLA models across diverse robotic manipulation scenarios. The main results are summarized in Table~\ref{tab:maniskill_results} and Table~\ref{tab:robotwin_results}.

\begin{table}[t]\centering
\caption{
Results on ManiSkill3 under in-distribution and out-of-distribution evaluation. 
Results are measured by success rate (\%). 
The upper part reports SFT results, while the lower part reports RL methods. 
Avg denotes the average performance across all OOD settings.
%
}
\label{tab:maniskill_results}
\vspace{-5pt}
\resizebox{1\linewidth}{!}{
\begin{tabular}{l|c|ccc|c}\toprule
\multirow{2}{*}{\textbf{Model}} & \multirow{2}{*}{\textbf{In-Distribution}} & \multicolumn{4}{c}{\textbf{Out-Of-Distribution}} \\
& & \textbf{Vision} & \textbf{Semantic} & \textbf{Execution} & \textbf{Avg} \\\midrule
$\pi_{0}$ + SFT &38.4 &32.6 &8.4 &13.2 &18.1 \\
$\pi_{0.5}$ + SFT &40.1 &40.2 &16.6 &22.4 &26.4 \\
OpenVLA-OFT + SFT &28.1 &27.7 &13.0 &11.7 &17.5 \\
\midrule
$\pi_{0}$ + Flow-SDE &78.8 &61.1 &25.4 &31.5 &39.3 \\
$\pi_{0}$ + Flow-Noise &77.8 &63.4 &23.1 &24.2 &36.9 \\
$\pi_{0.5}$ + Flow-SDE &90.9 &68.0 &34.5 &45.4 &49.3 \\
$\pi_{0.5}$ + Flow-Noise &89.7 &69.9 &35.5 &54.9 &53.4 \\
OpenVLA-OFT + GRPO &94.1 &84.7 &45.5 &44.7 &58.3 \\
OpenVLA-OFT + PPO &\textbf{97.7} &92.1 &64.8 &73.6 &76.8 \\
\textbf{OpenVLA-OFT + Feat2Go (Ours)}  &96.9 &\textbf{95.6} &\textbf{70.3} &\textbf{82.9} &\textbf{82.9} \\
\bottomrule
\end{tabular}}
\end{table}

\textbf{Single-Arm Manipulation on ManiSkill3\ \ } 
Table~\ref{tab:maniskill_results} presents the evaluation results on ManiSkill3. 
While fundamental SFT models exhibit reasonable capabilities on in-distribution tasks, they suffer severe performance degradation when exposed to OOD scenarios.
Although standard reinforcement learning paradigms alleviate this generalization gap, our Feat2Go approach yields state-of-the-art performance across all OOD zero-shot settings.
Specifically, OpenVLA-OFT + Feat2Go attains the highest success rates in Vision ($95.6\%$), Semantic ($70.3\%$), and Execution ($82.9\%$) evaluations, culminating in an impressive $82.9\%$ average OOD performance. 
This marks a remarkable $+65.4\%$ absolute improvement over the initial SFT weights. 
Furthermore, our method maintains highly competitive in-distribution proficiency ($96.9\%$), demonstrating that Feat2Go effectively prevents catastrophic forgetting of original skills while mapping nuanced visual features robustly for dynamic environments.

\begin{table}[htpb]
\centering
\caption{
Evaluation on six RoboTwin 2.0 tasks.
Results are measured by success rate (\%).
Avg represents the average success rate across all six tasks.
%
}
\label{tab:robotwin_results}
\vspace{-3pt}
\small
\setlength{\tabcolsep}{4pt}
\resizebox{\linewidth}{!}{%
\begin{tabular}{l c c c c c c c}
\toprule
\multirow{2}{*}{\textbf{Model}}
& \multirow{2}{*}{\textit{Handover Block}}
& \multirow{2}{*}{\textit{Lift Pot}}
& \multirow{2}{*}{\textit{Move Can Pot}}
& \textit{Pick Dual}
& \textit{Place Container}
& \textit{Place Empty}
& \multirow{2}{*}{\textbf{Avg}} \\
& & & & \textit{Bottles} & \textit{Plate} & \textit{Cup} & \\
\cmidrule(lr){1-8}
DP + SFT & 0.0 & 0.0 & 0.0 & 0.0 & 0.0 & 0.0 & 0.0 \\
ACT + SFT & 0.0 & 0.0 & 4.0 & 0.0 & 1.0 & 0.0 & 0.8 \\
DP3 + SFT & 0.0 & 0.0 & 6.0 & 1.0 & 1.0 & 1.0 & 1.5 \\
RDT + SFT & 14.0 & 9.0 & 12.0 & 13.0 & 17.0 & 7.0 & 12.0 \\
$\pi_{0}$ + SFT & 8.0 & 36.0 & 21.0 & 12.0 & 45.0 & 11.0 & 22.2 \\
OpenVLA-OFT + SFT & 28.1 & 3.1 & 9.4 & 20.3 & 54.7 & 75.8 & 31.9 \\\midrule
OpenVLA-OFT + SimpleVLA-RL & 57.8 & 64.1 & 61.2 & 68.3 & 82.1 & 94.2 & 71.3 \\
OpenVLA-OFT + GRPO & 71.9 & 71.9 & 83.6 & 90.6 & 93.0 & 93.8 & 84.1 \\
\textbf{OpenVLA-OFT + Feat2Go (Ours)} & \textbf{76.6} & \textbf{78.9} & \textbf{91.4} & \textbf{93.8} & \textbf{96.9} & \textbf{95.3} & \textbf{88.8} \\
\bottomrule
\end{tabular}%
}
\end{table}
\textbf{Bimanual Manipulation on RoboTwin 2.0\ \ } 
Table~\ref{tab:robotwin_results} presents the evaluation results on RoboTwin 2.0.
In bimanual tasks, conventional behavior cloning baselines (DP, ACT, DP3) collapse almost entirely (averaging $\le 1.5\%$), and robust foundation models like $\pi_{0}$ + SFT only reach $22.2\%$. 
In contrast, RL-aligned VLAs showcase significant improvements. Among them, our Feat2Go strategy consistently surpasses other advanced RL baselines like SimpleVLA-RL and GRPO.
Feat2Go establishes new state-of-the-art results across all six challenging bimanual tasks, recording an average success rate of $88.8\%$.
Notably, Feat2Go excels in tasks requiring high-precision spatial synchronization; for example, it completely outperforms other architectures on \textit{Place Container Plate} ($96.9\%$) and \textit{Move Can Pot} ($91.4\%$). 
Overall, our method yields an extraordinary $+56.9\%$ average absolute improvement ($\Delta$) over the base OpenVLA-OFT + SFT model (with a peak improvement of $+82.0\%$ on \textit{Move Can Pot}), verifying that targeted embodied value representation inherently accelerates and stabilizes policy learning for complex bimanual coordination.
\subsection{Ablation Studies}

\begin{table}[t]\centering
\caption{Ablation study comparing our Feat2Go with Steps-To-Go on ManiSkill3.}
\label{tab:steps_to_go_results}
\resizebox{1\linewidth}{!}{
\begin{tabular}{l|c|ccc|c}\toprule
\multirow{2}{*}{\textbf{Model}} & \multirow{2}{*}{\textbf{In-Distribution}} & \multicolumn{4}{c}{\textbf{Out-Of-Distribution}} \\
& & \textbf{Vision} & \textbf{Semantic} & \textbf{Execution} & \textbf{Avg} \\\midrule
OpenVLA-OFT + SFT & 28.1 & 27.7 & 13.0 & 11.7 & 17.5 \\
OpenVLA-OFT + Steps-To-Go & 95.7 & 93.5 & 67.0 & 76.4 & 79.0 \\
\textbf{OpenVLA-OFT + Feat2Go (Ours)}  & \textbf{96.9} & \textbf{95.6} & \textbf{70.3} & \textbf{82.9} & \textbf{82.9} \\
\bottomrule
\end{tabular} }
\vspace{-15pt}
\end{table}

\begin{wrapfigure}{R}{0.5\textwidth}
    \vspace{-15pt}
    \centering
    \includegraphics[width=\linewidth]{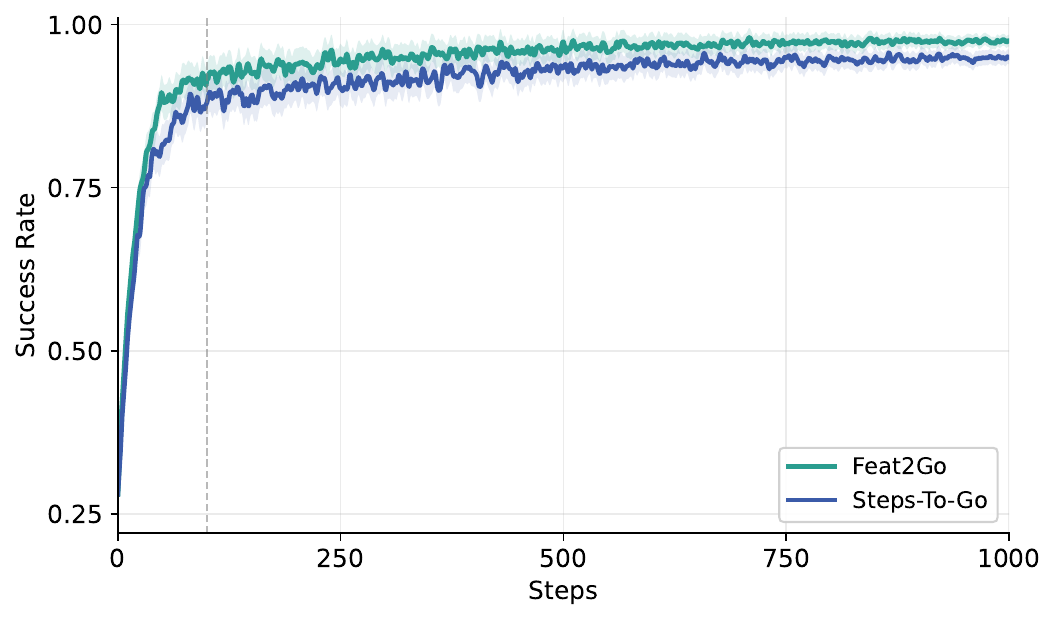}
    \caption{Training curves of the reinforcement learning phase using Feat2Go and Steps-To-Go as value estimation targets.}
    \label{fig:ablation_f2g_vs_steps_to_go}
    \vspace{-10pt}
\end{wrapfigure}

To validate the core intuition behind our framework, we ablate the proposed Feat2Go representation against a direct Steps-To-Go baseline~\citep{ghasemipour2025selfimproving}. 
In the latter paradigm, the objective is constrained to the naive regression of the temporal distance to task completion rather than implicitly modeling the generalized state-value manifold.
When deployed as the value estimation target during the reinforcement learning phase, the purely temporal distance metric of Steps-To-Go is heavily susceptible to rollout variance in complex scenarios. 
In contrast, Feat2Go offers a continuous, semantics-aware state-value embedding that grants the VLA model a more robust capability to capture intricate task dynamics.
The evaluation outcomes summarized in Table~\ref{tab:steps_to_go_results} substantiate this underlying hypothesis. 
While initializing with the raw Steps-To-Go representation provides a measurable foundation, Feat2Go uniformly dominates across every generalization axis. 
Notably, in the rigorous \textit{Execution} evaluation, which necessitates intense spatial robustness against dynamic task setups, Feat2Go delivers a substantial absolute improvement of $+6.5\%$. 
Ultimately, these results confirm that distilling the embodied task progress via Feat2Go rather than employing blunt temporal scalars endows the policy with a structurally richer inductive bias, leading to more resilient zero-shot performance in OOD conditions.

\section{Conclusion}
\label{sec:conclusion}
In this work, we introduced Feat2Go, a novel representation framework explicitly designed to be integrated with reinforcement learning to enhance the robust alignment and generalization of VLA models. 
Unlike traditional approaches that rely on sparse rewards or blunt temporal metrics, our method distills embodied task progress into a continuous, semantics-aware state-value embedding. 
By serving as a dense and highly informative value estimation target during the RL training phase, this structurally richer inductive bias effectively mitigates rollout variance and guides the policy toward superior task execution capabilities.
Extensive evaluations across the ManiSkill3 and RoboTwin 2.0 benchmarks demonstrated that RL algorithms augmented with Feat2Go establish new state-of-the-art performance in rigorous visual, semantic, and spatial conditions. 
Crucially, our framework underscores a promising paradigm shift: efficiently translating pre-trained representation manifolds into grounded, dynamic reinforcement signals, empowering embodied agents to break through traditional RL performance bottlenecks without being misled by inaccurate or suboptimal reward formulations.
While Feat2Go demonstrates strong simulation results, future work will focus on scalable and safe real-world data collection to address hardware constraints and deployment costs.

\bibliographystyle{plainnat}
\bibliography{reference}

\end{document}